\documentclass[letterpaper, 10 pt, conference]{ieeeconf}  

\IEEEoverridecommandlockouts                              

\usepackage{graphicx}
\usepackage{booktabs}
\usepackage[table]{xcolor}
\usepackage{array}
\usepackage{makecell}
\usepackage{multirow}
\usepackage{tabularx}
\usepackage{adjustbox}
\usepackage{siunitx}
\usepackage{placeins}
\usepackage{float}
\usepackage{balance}
\usepackage[normalem]{ulem} 
\usepackage{amsmath,amssymb}
\DeclareMathSizes{7.4}{7.4}{5}{5}
\DeclareMathSizes{7.5}{7.5}{5}{5}

\title{\LARGE \bf
ViLoMan: Learning \uline{Vi}sual--Proprioceptive Whole-Body \uline{Lo}co-\uline{Man}ipulation Skills for Humanoid Robots
}

\author{Zejie Tian$^{1,2,3}$, Ruibing Hou$^{1,*}$, Bingpeng Ma$^{2}$,
B{\"o}rje F. Karlsson$^{3}$, and Shiguang Shan$^{1,2}$%
\thanks{$^{1}$State Key Laboratory of AI Safety, Institute of Computing Technology, CAS, China.}%
\thanks{$^{2}$University of Chinese Academy of Sciences (CAS), China.}%
\thanks{$^{3}$Beijing Academy of Artificial Intelligence (BAAI).}%
\thanks{$^{*}$Corresponding author: Ruibing Hou (\texttt{houruibing@ict.ac.cn}).}%
}
\begin{document}

\hbadness=10000
\vbadness=10000

\maketitle
\thispagestyle{empty}
\pagestyle{empty}

\begin{abstract}
Humanoid loco-manipulation requires adaptive whole-body coordination to seamlessly integrate locomotion and physical interaction. Despite recent advances, learning autonomous loco-manipulation remains challenging due to the scarcity of diverse, physically executable robot--object interaction data and the difficulty of learning unified whole-body control directly from onboard observations. We present \textbf{ViLoMan}, a scalable framework for autonomous humanoid loco-manipulation. ViLoMan first transforms partial kinematic demonstrations of human--object interactions into complete, physically executable robot trajectories. It then leverages these trajectories within a teacher--student distillation framework to learn a unified policy that maps egocentric depth observations and proprioceptive measurements directly to joint-level whole-body actions. During deployment, the policy requires neither reference motions nor intermediate commands. We evaluate ViLoMan on door-closing tasks across diverse door configurations and robot initial conditions in both simulation and the real world. Experimental results demonstrate that a single policy enables a Unitree G1 humanoid to complete the full task using only onboard depth sensing and proprioception, while generalizing robustly across task variations and transferring effectively from simulation to reality. Project page: \texttt{viloman-anonymous.pages.dev}.
\end{abstract}

\providecommand{\cmark}{\ensuremath{\checkmark}}
\providecommand{\xmark}{\ensuremath{\times}}

\begin{table*}[!t]
  \centering
  \caption{Comparison of representative humanoid loco-manipulation methods.}
  \label{tab:physical_execution_comparison}

  \setlength{\tabcolsep}{4pt}
  \renewcommand{\arraystretch}{1.08}
  \footnotesize

  \begin{tabular*}{\textwidth}{@{\extracolsep{\fill}}lccccc@{}}
    \toprule
    \textbf{Method}
    & \textbf{Onboard Vision}
    & \textbf{Multi-Motion Policy}
    & \textbf{Unified WBC}
    & \textbf{No Ref. Motion}
    & \textbf{No Interm. Cmd.} \\
    \midrule

    HDMI~\cite{weng2025hdmi}
    & \xmark & \xmark & \cmark & \xmark & \cmark \\

    ResMimic~\cite{zhao2025resmimic}
    & \xmark & \xmark & \cmark & \xmark & \cmark \\

    SUGAR~\cite{wu2026sugar}
    & \xmark & \cmark & \cmark & \cmark & \xmark \\

    VisualMimic~\cite{yin2025visualmimic}
    & \cmark & \cmark & \cmark & \cmark & \xmark \\

    ULTRA~\cite{ultra2026}
    & \cmark & \cmark & \cmark & \cmark & \xmark \\

    Imagine2Real~\cite{imagine2real2026}
    & \xmark & \cmark & \cmark & \xmark & \xmark \\

    OmniContact~\cite{yu2026omnicontact}
    & \xmark & \cmark & \cmark & \cmark & \xmark \\

    VIRAL~\cite{he2026viral}
    & \cmark & \cmark & \xmark & \cmark & \xmark \\

    DoorMan~\cite{xue2025doorman}
    & \cmark & \cmark & \xmark & \cmark & \xmark \\

    \midrule
    \textbf{ViLoMan (ours)}
    & \cmark & \cmark & \cmark & \cmark & \cmark \\

    \bottomrule
  \end{tabular*}

  \vspace{1.5mm}
  \begin{minipage}{0.99\textwidth}
    \scriptsize
    \raggedright
    \textbf{Onboard Vision:} 
    onboard visual observations available during deployment;
\textbf{Multi-Motion Policy:} 
a single policy capable of executing multiple interaction motions; 
\textbf{Unified whole-body control (WBC):} joint upper- and lower-body control without decoupled locomotion and manipulation controllers;
\textbf{No Ref.\ Motion:} no external or precomputed motion reference required at runtime;
\textbf{No Interm.\ Cmd.:} no intermediate control command required at runtime.
  \end{minipage}
\end{table*}

\section{INTRODUCTION}
\label{sec:introduction}

Humanoid loco-manipulation requires tightly coordinated control of locomotion, balance, and manipulation during physical interaction with the environment. Recent advances have enabled humanoid robots to perform increasingly complex and contact-rich loco-manipulation skills~\cite{zhang2025falcon,yu2026omnicontact,lee2025stageact,xue2025doorman}, representing substantial progress toward dynamic whole-body behaviors. However, generalizing such capabilities across diverse real-world scenarios remains challenging.

A fundamental challenge lies in acquiring diverse and physically executable loco-manipulation data at scale. Collecting whole-body interaction data directly on humanoid robots, whether through teleoperation or reinforcement learning, is costly and often task-specific~\cite{ze2025twist2,yu2026oasis,he2026viral,xue2025doorman}. Human--object interaction (HOI) datasets provide abundant demonstrations of whole-body physical interactions~\cite{li2023omomo,jiang2024trumans}; however, directly retargeting human motions to humanoid robots may violate robot-specific morphological and contact constraints. Moreover, these demonstrations typically capture only the interaction phase, omitting the approach behaviors required to construct complete trajectories from diverse initial robot configurations. Therefore, leveraging HOI data requires transforming partial human demonstrations into complete and physically executable robot trajectories.

A second challenge is learning autonomous whole-body control directly from onboard observations. The high-dimensional coordination of locomotion, balance, and manipulation 
makes direct vision-to-action learning difficult. Existing methods reduce this complexity by relying on motion references~\cite{weng2025hdmi,zhao2025resmimic,imagine2real2026}, introducing intermediate representations such as sparse keypoints and velocity commands~\cite{yin2025visualmimic,wu2026sugar,yu2026omnicontact,li2026vaic}, or decoupling locomotion and manipulation into separate controllers~\cite{he2026viral,xue2025doorman}. 
However, these designs either depend on externally specified guidance or impose predefined control structure, potentially limiting adaptive whole-body coordination.
As summarized in Table~\ref{tab:physical_execution_comparison}, no prior approach simultaneously supports onboard vision-based control, unified whole-body execution, and deployment without motion references or intermediate commands.

\begin{figure}[t]
  \centering
  \includegraphics[width=0.95\columnwidth]{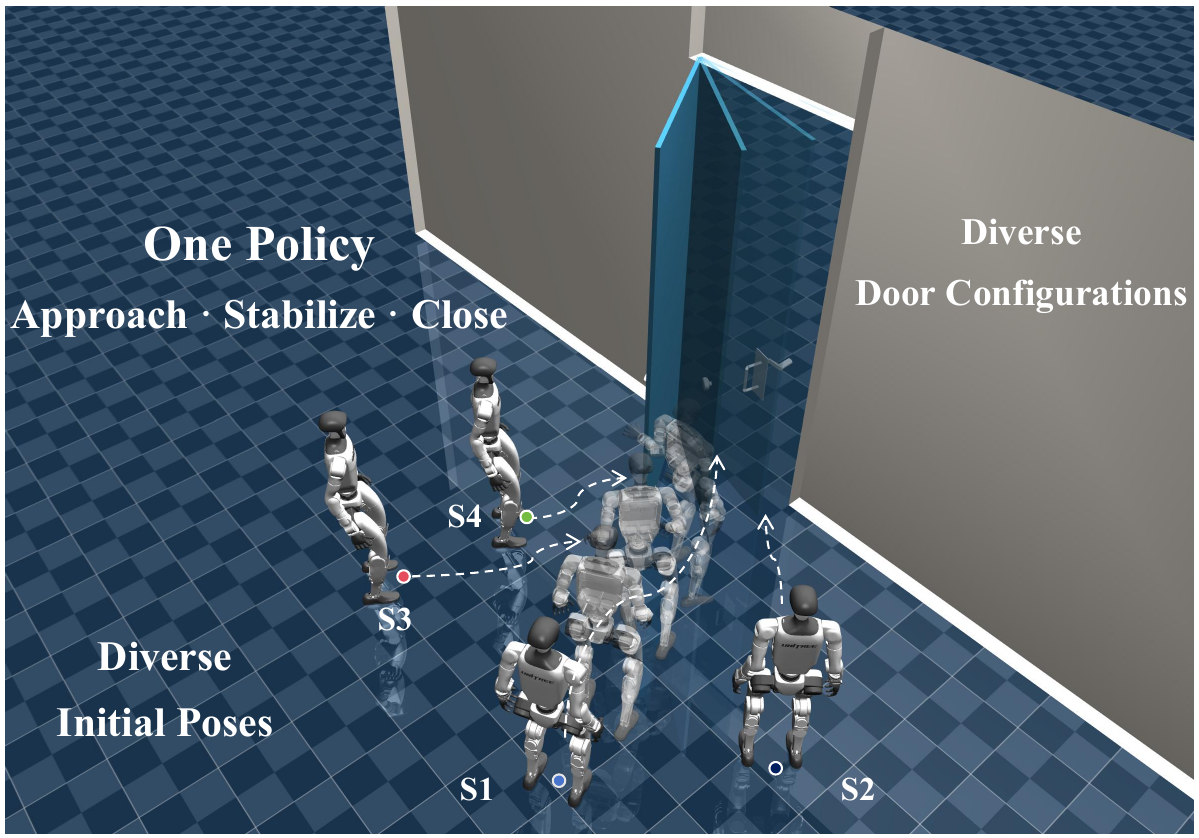}
  \caption{ViLoMan enables a single policy to perform closed-loop door closing across diverse initial robot poses, door configurations, and interaction conditions, covering approach, stabilization, and closing.}
  \label{fig:teaser}
\end{figure}

To address these challenges, we propose ViLoMan, a scalable framework for autonomous humanoid loco-manipulation. ViLoMan first constructs diverse, complete motion references by augmenting retargeted HOI motions with generated approach motions initialized from diverse robot configurations. It then transforms these kinematic references into physically executable behaviors through physics-based motion tracking that accounts for both humanoid dynamics and robot--object interactions. In this way, ViLoMan systematically converts abundant human motion data into scalable robot experience, eliminating the need to collect task-specific trajectories on physical robots.

Leveraging this experience, ViLoMan learns a unified closed-loop policy that directly maps egocentric depth observations and proprioception to whole-body actions, jointly coordinating locomotion, balance, and manipulation. To make direct vision-to-action learning tractable, we first train a \textit{residual} teacher policy conditioned on motion references and privileged object states to acquire complex interaction behaviors. We then distill these behaviors into the visual--proprioceptive policy using Dataset Aggregation (DAgger)~\cite{ross2011dagger}. At deployment, the resulting policy operates solely on onboard observations, without requiring motion references or intermediate control commands.

We evaluate ViLoMan on door closing, a representative loco-manipulation task that requires the humanoid to approach the door, establish contact, and maintain coordinated whole-body interaction. Our evaluation spans diverse door configurations, initial door angles, and robot starting positions, requiring the policy to continuously adapt to varying interaction conditions. Experiments in both simulation and on a real Unitree G1 demonstrate that a single ViLoMan policy can execute the complete loco-manipulation sequence using only onboard depth observations and proprioception. 

\section{RELATED WORK}

\noindent
\textbf{Human Motion Data for Humanoid Loco-Manipulation.} \ 
Existing structured 3D interaction datasets cover diverse scenarios, including human--object interactions, human--scene interactions, and multi-object manipulation. These datasets typically pair parametric full-body motion~\cite{pavlakos2019smplx} with synchronized object trajectories, providing rich motion and contact priors for humanoid policy learning~\cite{li2023omomo,jiang2024trumans,lu2025humoto}. To bridge the morphological gap between humans and humanoid robots, general motion-retargeting methods preserve motion geometry while ensuring robot trackability~\cite{araujo2025gmr}. Interaction-aware methods additionally preserve human--object relationships, extending retargeting from human motion alone to the joint transfer of human motion, object motion, and contacts~\cite{yang2025omniretarget}. Beyond structured 3D datasets, recent studies derive interaction supervision from videos. HDMI and SUGAR extract human--object trajectories and contacts from unstructured videos and transform them into executable robot skills through physics-based policy learning~\cite{weng2025hdmi,wu2026sugar}. GRAIL, Imagine2Real, and ExoActor instead generate task-conditioned interaction videos and translate them into humanoid behaviors through metric 4D reconstruction, sparse keypoint trajectories, or human motion  estimation~\cite{xie2026grail,imagine2real2026,zhou2026exoactor}. Although these approaches have substantially broadened the available sources of interaction data, constructing loco-manipulation datasets that simultaneously ensure physical executability, interaction fidelity, and complete long-horizon trajectories remains challenging.

\noindent
\textbf{Humanoid Whole-Body Loco-Manipulation.} \ 
One line of research extends motion imitation and reference-conditioned control to contact-rich loco-manipulation. These methods jointly track humanoid and object motions, adapt motion priors, 
or explicitly model interaction forces, object dynamics, upper-body compliance, and contact information~\cite{weng2025hdmi,zhao2025resmimic,wu2026sugar,zhang2025falcon,gentlehumanoid2025,chen2026scenebot}.
Other approaches replace full-body references with compact task representations or objectives, such as velocity commands, sparse keypoints, contact flows, and multimodal objectives. These formulations improve skill reuse, long-horizon composition, and generalization across object geometries~\cite{li2026vaic,imagine2real2026,yu2026omnicontact,ultra2026}.
More recently, closed-loop vision-based approaches have combined onboard RGB or depth observations with proprioception, using motion supervision and reinforcement learning to enable perception-driven execution~\cite{yin2025visualmimic,he2026viral,xue2025doorman,wu2026php}. Nevertheless, many existing systems still rely on motion references or structured task-space commands throughout execution, while others employ hierarchical control architectures or explicitly decouple locomotion from manipulation. Consequently, unified vision-based joint-level control for contact-rich whole-body tasks remains largely underexplored.

\begin{figure*}[!t]
  \centering
  \includegraphics[width=0.9\textwidth]{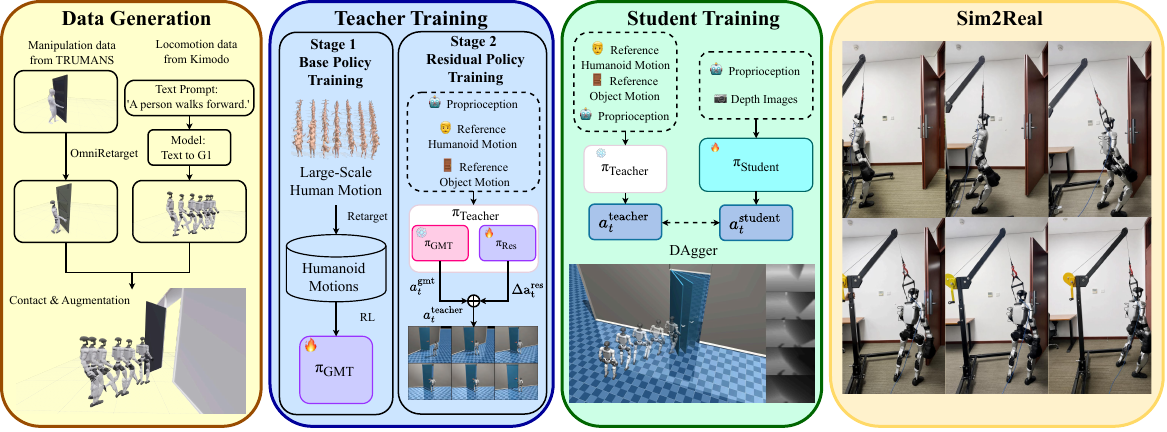}
  \caption{ViLoMan training and deployment pipeline. Human interaction motions are retargeted to the humanoid, augmented with generated approach motions, and tracked by a general motion-tracking prior with residual interaction policies. Motion-clustered specialist policies collectively form a privileged teacher, which supervises a visual--proprioceptive student on states encountered during student rollouts. At deployment, the student maps onboard observations directly to 29-dimensional joint-position actions, without requiring motion references or intermediate commands.}
  \label{fig:pipeline}
\end{figure*}

\section{METHOD}
\label{sec:method}

\subsection{Overview}
\label{sec:method_overview}

\noindent
\textbf{Problem setting.} \ We control a 29-DoF Unitree G1 humanoid at 50\,Hz. At each time step $t$, the deployed student policy receives a proprioceptive history $s_t^{\mathrm{prop}}\in\mathbb{R}^{808}$, comprising body-frame angular velocity, projected gravity, joint positions and velocities, and previous actions, along with four consecutive egocentric depth images $I_{t-3:t}^{\mathrm{depth}}\in\mathbb{R}^{4\times36\times64}$. The simulated camera is calibrated to match the viewpoint of the G1's head-mounted depth camera. Based on these observations, the policy $\pi_{\mathrm{Student}}$ outputs a 29-dimensional joint-position action $a_t^{\mathrm{student}}\in\mathbb{R}^{29}$.
Our objective is to learn a closed-loop visuomotor policy for long-horizon loco-manipulation that operates solely on onboard observations at deployment, without access to motion references or privileged object states.

\noindent
\textbf{Method Overview.} \ 
As illustrated in Fig.~\ref{fig:pipeline}, our framework consists of three stages. First, we construct complete loco-manipulation sequences by augmenting retargeted human interactions with diverse approach trajectories, followed by physics-aware refinement. Second, we equip a general motion-tracking policy with \textit{residual interaction adaptation} to learn a reference-conditioned teacher. Finally, we distill the privileged teacher into a visuomotor policy via DAgger, transferring long-horizon behaviors learned from privileged states and motion references to a policy driven by onboard visual observations.

\subsection{Physically Executable Loco-Manipulation Data}
\label{sec:reference_construction}
Human interaction data capture diverse manipulation behaviors, but transferring these behaviors to humanoid loco-manipulation requires bridging three key gaps: embodiment differences, missing approach motions, and the physical infeasibility of kinematic references. To bridge these gaps, we construct a dataset of complete, physically executable loco-manipulation motions through the following stages.

\subsubsection{Interaction-Preserving Retargeting}
We use human--door closing interactions from TRUMANS~\cite{jiang2024trumans}, represented using SMPL-X~\cite{pavlakos2019smplx}, as source motions. To account for embodiment differences while preserving task-critical human--object geometry, we retarget the human motions to the humanoid using OmniRetarget~\cite{yang2025omniretarget}. Synchronizing the corresponding door trajectories with the retargeted motions yields coupled humanoid--object interaction segments. 

\subsubsection{Generative Approach Augmentation}
TRUMANS clips typically begin near the door and lack approach motions. We use Kimodo~\cite{rempe2026kimodo}, a G1-compatible motion generator, to generate waypoint-conditioned approach trajectories from diverse initial poses to the corresponding interaction-ready pose. Each approach trajectory is temporally aligned and smoothly concatenated with its interaction segment, yielding complete loco-manipulation motions. Sampling 10 approaches for each of 71 interaction segments produces 710 complete trajectories.

\subsubsection{Door Augmentation}
\begin{figure}[!t]
    \centering
    \includegraphics[width=0.8\columnwidth]{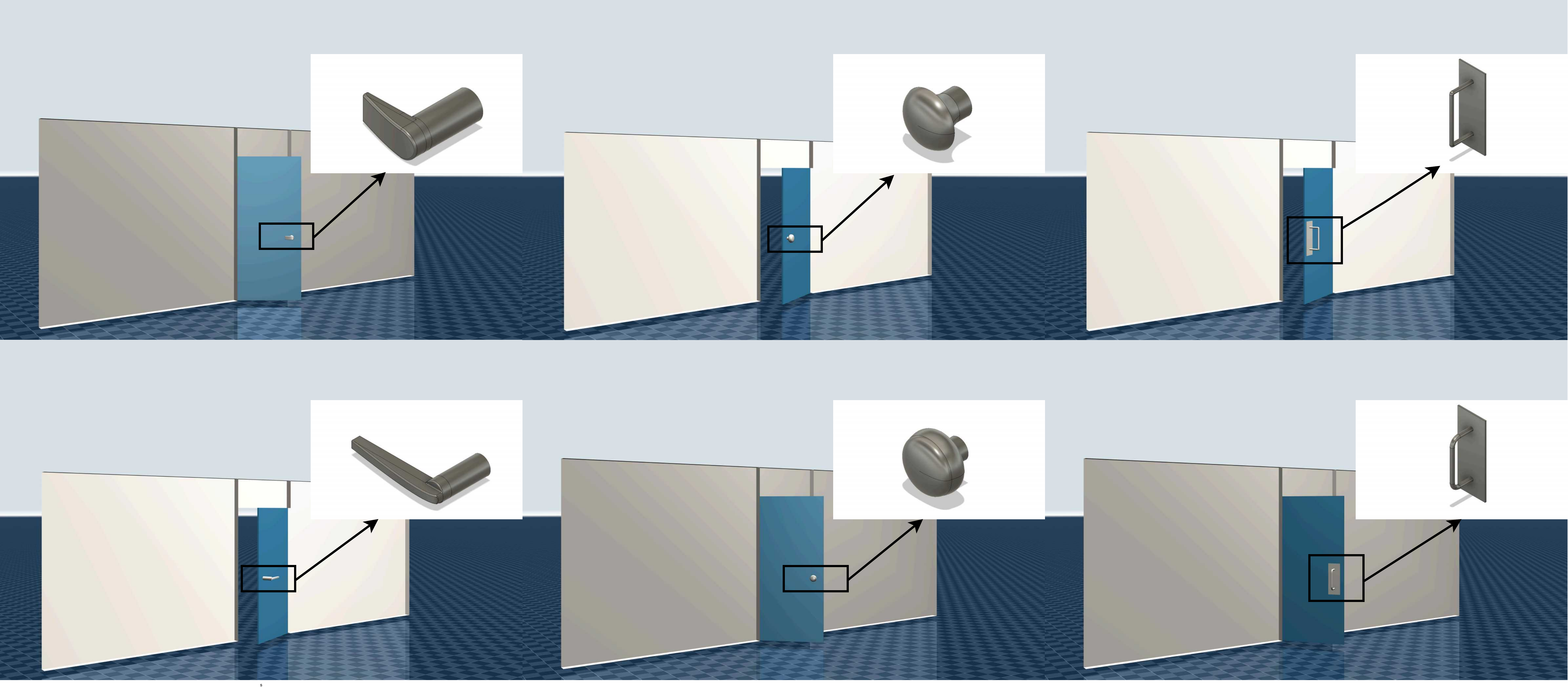}
    \caption{Parameterized DoorGym assets used to diversify interaction geometry. The six configurations cover lever, round, and pull handles on left- and right-hinged doors.}
    \label{fig:doorgym}
\end{figure}

We further diversify these 710 trajectories using parameterized  door assets from DoorGym~\cite{urakami2019doorgym}. For each trajectory, we randomly instantiate a door by varying its handle type (lever, pull, or round), hinge side, initial opening angle, and physical dimensions, while preserving the intended humanoid--door interaction geometry, as illustrated in Fig.~\ref{fig:doorgym}. 

\subsubsection{Physics-Aware Refinement}
The augmented trajectories remain purely kinematic and may violate system dynamics or contact constraints. For each humanoid--door reference trajectory, we train a dedicated policy to jointly track the humanoid and door motions in simulation. The resulting rollouts adapt the humanoid's posture, balance, and contact behavior to satisfy physical constraints while preserving the intended interaction. We retain the most feasible rollout as the refined trajectory.

\subsubsection{Filtering}

We remove redundant trajectories and discard rollouts exhibiting tracking failures or noticeable physical artifacts. This filtering process yields 409 training trajectories and 106 test trajectories. We split the data according to the source TRUMANS clips to prevent overlap between the training and test sets. Both splits encompass diverse interaction behaviors, handle types, door angles, and initial robot configurations, as summarized in Table~\ref{tab:dataset_diversity}.

\subsection{Reference-Conditioned Interaction Teacher}
\label{sec:teacher}
Our objective is to learn a robust interaction teacher that tracks diverse reference humanoid--object trajectories and provides reliable action supervision for subsequent student policy distillation.  Learning such a controller directly from interaction demonstrations is challenging, because it requires both general whole-body control and adaptation to interaction-specific dynamics. We therefore initialize the teacher with a pretrained \emph{general motion-tracking (GMT) policy} and introduce a \emph{privileged residual policy} for interaction-specific adaptation.

\subsubsection{General Motion Tracking Prior}
We adopt the GMT-based whole-body motion tracker from GentleHumanoid~\cite{gentlehumanoid2025} as the base policy. It observes  $o_t^{\mathrm{GMT}}$, comprising the robot’s proprioceptive history, multi-step reference features for the root and joints, and boot/compliance indicators. The pretrained GMT policy produces a nominal action $a_t^{\mathrm{gmt}} \in \mathbb{R}^{29}$:
\begin{equation}
a_t^{\mathrm{gmt}}=\pi_{\mathrm{GMT}}\left(o_t^{\mathrm{GMT}}\right).
\end{equation}
This action provides a task-agnostic prior for humanoid locomotion and whole-body motion tracking.

\subsubsection{Residual Interaction Adaptation}
Motion tracking alone does not ensure successful object interaction, which depends on contact transitions and coupled humanoid--object dynamics. We therefore introduce a privileged residual policy $\pi_{\mathrm{Res}}$ to adapt the GMT prior to interaction-specific dynamics. The residual policy is conditioned on an interaction context \(c_t^{\mathrm{int}}\) encoding the task stage, door state, contact target, and reference hinge motion. The resulting teacher observation is $o_t^{\mathrm{teacher}}=\left[o_t^{\mathrm{GMT}},c_t^{\mathrm{int}}\right]$. 
During training, $\pi_{\mathrm{GMT}}$ remains frozen, while \(\pi_{\mathrm{Res}}\) predicts an interaction-dependent correction that is added to the nominal action:
\begin{equation}
a_t^{\mathrm{teacher}}=\pi_{\mathrm{Teacher}}\left(o_t^{\mathrm{teacher}}\right)=a_t^{\mathrm{gmt}}+\pi_{\mathrm{Res}}\left(o_t^{\mathrm{teacher}}\right). 
\end{equation}
This residual setting enables the teacher policy to learn interaction-specific corrections without relearning general whole-body control.

\subsubsection{Teacher Training}
We optimize the residual policy using Proximal Policy Optimization (PPO)~\cite{schulman2017ppo} with stage-conditioned rewards. The rewards emphasize locomotion and humanoid tracking during approach, stabilization and contact preparation before contact, and humanoid--object trajectory tracking and contact consistency during door closing. Reward terms and stage-specific weights are listed in Table~\ref{tab:privileged_teacher_reward}. We randomize robot dynamics, contact and ground parameters, door dynamics, and external perturbations to improve robustness and sim-to-real transfer (Table~\ref{tab:domain_randomization}). To model diverse interactions, we cluster demonstrations by motion characteristics and train one residual specialist per cluster. These specialists form the privileged interaction teacher and supervise DAgger-based distillation of the visual--proprioceptive student policy.

\subsection{Reference-Free Visuomotor Distillation}
\label{sec:student}

Although the privileged interaction teacher achieves robust long-horizon performance, it relies on future reference motions and privileged object states unavailable at deployment. We therefore distill the teacher into a reference-free visuomotor policy that enables humanoid--object interaction using only onboard sensory observations. Specifically, the student policy $\pi_{\mathrm{Student}}$ receives the observation $o_t^{\mathrm{student}} = [s_t^{\mathrm{prop}},I_{t-3:t}^{\mathrm{depth}}]$, where $s_t^{\mathrm{prop}}$ and $I_{t-3:t}^{\mathrm{depth}}$ are defined in Section \ref{sec:method_overview}. 
To enhance robustness to visual-domain variations, we randomize the Gaussian blur applied to the simulated depth observations. 
A residual convolutional neural network (CNN) encodes the depth observations into a 32-dimensional  feature that is fused with proprioception to predict the student action:
\begin{equation}
a_t^{\mathrm{student}}=\pi_{\mathrm{Student}}(o_t^{\mathrm{student}}).
\end{equation}

Direct offline behavior cloning (BC) is susceptible to covariate shift: small prediction errors can accumulate over time and drive the student into states absent from the demonstration data. To mitigate this issue, we employ online DAgger~\cite{ross2011dagger}, which continually adapts the training distribution to the state distribution induced by the current student policy. 
For each simulator state $\xi_t\sim d_{\pi_{\mathrm{Student}}}$ encountered during a student rollout, the corresponding student observation $o_t^{\mathrm{student}}$ and teacher observation $o_t^{\mathrm{teacher}}$ are constructed from the same underlying simulator state $\xi_t$, and the teacher provides the supervisory action. The collected batch is therefore
\begin{equation}
\mathcal{B}=\left\{
\left(o_t^{\mathrm{student}},a_t^{\mathrm{teacher}}\right)
\;\middle|\;
\begin{aligned}
\xi_t &\sim d_{\pi_{\mathrm{Student}}},\\
a_t^{\mathrm{teacher}}
&=\pi_{\mathrm{Teacher}}(o_t^{\mathrm{teacher}})
\end{aligned}
\right\}.
\end{equation}
Each training minibatch follows $\mu$, with two-thirds of its samples drawn from the current batch $\mathcal{B}$ and one-third from the pre-update replay buffer. 
The current batch enters the bounded replay buffer after the student update.
After normalizing the student and teacher actions per dimension to form $\tilde{a}^{\mathrm{student}}$ and $\tilde{a}^{\mathrm{teacher}}$, we minimize
\begin{equation}
\begin{aligned}
\mathcal{L}_{\mathrm{DAgger}}
&=\mathbb{E}_{(o^{\mathrm{student}},a^{\mathrm{teacher}})\sim\mu}
\bigl[
\bigl\|\tilde a^{\mathrm{student}}
-\tilde a^{\mathrm{teacher}}\bigr\|_2^2
\bigr].
\end{aligned}
\end{equation}
By iteratively aggregating data under the evolving student-induced state distribution, the student progressively recovers the teacher’s interaction capabilities while removing its dependence on privileged information and reference trajectories. This procedure ultimately yields a deployable, reference-free visuomotor policy.


\section{EXPERIMENTS}
Our evaluation is designed to answer three questions:
(Q1) Are the privileged interaction teacher and online distillation necessary?
(Q2) How does direct joint-level control compare with intermediate-level execution interfaces?
(Q3) Can the learned policy adapt beyond reference-motion replay and transfer effectively to a physical robot?

\subsection{Experimental Setup and Metrics}
All simulation experiments are conducted in the GPU-parallel mjlab environment~\cite{zakka2026mjlab}. The complete student-training setup uses 16,384 parallel simulation environments.
The policy is trained using the 409 trajectories constructed in the previous section and evaluated on a test set of 106 trajectories. Both splits span diverse handle categories, initial door angles, door geometries, and relative robot--door configurations. Detailed statistics are provided in Table~\ref{tab:dataset_diversity}.
For evaluation, each trajectory is executed in five randomized trials. In each trial, the domain-randomization parameters are sampled independently to assess policy robustness under diverse conditions. The success and survival rates are aggregated across all trials.

We use \textbf{survival rate} and \textbf{Door-Closing Success Rate (SR)} as the evaluation metrics. A rollout is counted as surviving if the robot remains physically stable and does not trigger early termination throughout the evaluation horizon. A rollout is considered successful only if it meets the survival criterion and reduces the door opening angle to below $10^\circ$ by the end of the episode. Reporting both metrics allows us to distinguish stability failures from task-execution failures.

\subsection{Teacher Adaptation and Online Distillation}

\begin{table}[t]
\caption{Policy comparison and ablation results on the training split and source-disjoint test splits.}
\label{tab:main_results}
\centering
\scriptsize
\setlength{\tabcolsep}{2.2pt}
\renewcommand{\arraystretch}{1.02}
\begin{tabular}{@{}lcccc@{}}
\toprule
& \multicolumn{2}{c}{\textbf{Door-Closing SR (\%) $\uparrow$}} & \multicolumn{2}{c}{\textbf{Survival (\%) $\uparrow$}} \\
\cmidrule(lr){2-3}\cmidrule(lr){4-5}
\textbf{Method} & \textbf{Train} & \textbf{Test} & \textbf{Train} & \textbf{Test} \\
\midrule
\multicolumn{5}{@{}l}{\textit{Privileged policies}} \\
\hspace{0.6em}GMT prior & 7.87 & 4.53 & \textbf{99.95} & \textbf{100.00} \\
\hspace{0.6em}GMT teacher (task data) & 80.64 & 73.96 & 89.68 & 88.49 \\
\hspace{0.6em}Teacher w/o physics-aware refinement & 79.85 & 76.98 & 98.04 & 96.79 \\
\cmidrule(lr){1-5}
\hspace{0.6em}\textbf{GMT with residual adaptation} & \textbf{88.12} & \textbf{84.91} & \underline{98.53} & \underline{98.11} \\
\midrule
\multicolumn{5}{@{}l}{\textit{Deployable students}} \\
\hspace{0.6em}1/4 training envs & 66.50 & 67.17 & 98.39 & 99.25 \\
\hspace{0.6em}1/2 training envs & 72.81 & 69.81 & 98.92 & 99.43 \\
\hspace{0.6em}CVAE & 74.91 & 76.42 & 99.07 & 98.68 \\
\hspace{0.6em}MLP-2 & 73.69 & 73.58 & 98.83 & 98.30 \\
\hspace{0.6em}MLP-4 & 76.77 & 73.77 & 99.22 & 99.43 \\
\hspace{0.6em}w/o vision & 43.72 & 33.96 & 94.62 & 94.72 \\
\hspace{0.6em}BC only & 36.82 & 31.89 & 57.51 & 56.98 \\
\cmidrule(lr){1-5}
\hspace{0.6em}\textbf{ViLoMan (MLP-3)} & \textbf{77.16} & \textbf{77.55} & \textbf{99.22} & \textbf{99.81} \\
\bottomrule

\end{tabular}
\end{table}

To address Q1, we ablate privileged teacher adaptation, physics-aware refinement, and online DAgger distillation. We also examine the effects of student architecture, visual input, and training scale in Table~\ref{tab:main_results}. During privileged-teacher training, episodes are initialized at the beginning of Stages 0, 1, and 2 with probabilities 0.25, 0.15, and 0.60, respectively.

As shown in Table~\ref{tab:main_results}, the generic motion tracker from GentleHumanoid~\cite{gentlehumanoid2025} achieves only 4.53\% Door-Closing SR, indicating that generic motion priors do not transfer zero-shot to contact-rich door-closing tasks. Training a task-specific GMT policy from scratch on our reference corpus increases the SR to 73.96\%, but yields a lower survival rate due to unstable motions and contacts. Residual interaction adaptation further improves the SR and survival rate to 84.91\% and 98.11\%, respectively, demonstrating the value of task-specific corrections for contact and door dynamics.  Removing physics-aware refinement reduces the test SR by 7.93 percentage points, confirming that foot--ground consistency, initial-pose refinement, and physically consistent long-horizon references facilitate stable interaction.

As shown in Table~\ref{tab:main_results}, offline behavior cloning fails to reproduce the teacher's closed-loop behavior because compounding errors drive the student beyond the demonstration distribution. Online DAgger addresses this covariate shift by providing corrective actions at student-visited states, improving the SR and survival by 45.66 and 42.83 percentage points over behavior cloning, respectively. These results demonstrate the importance of corrective supervision beyond nominal offline demonstrations.

Additional ablations examine the student architecture, visual observations, and online training scale. The three-layer MLP outperforms two- and four-layer MLPs, as well as the conditional variational autoencoder (CVAE). Removing visual observations  substantially  degrades performance, indicating that proprioception is insufficient to capture the door state and robot--door spatial relationship. Reducing the number of parallel training environments by one half and three quarters decreases test SR by 7.74 and 10.38 percentage points, respectively. These results support the three-layer MLP design and highlight the importance of visual feedback and sufficient online data aggregation for robust closed-loop interaction.

\begin{table}[t]
\caption{Comparison between direct joint-level control and intermediate-level execution interfaces. Execution time is averaged over successful trials.}
\label{tab:handoff_comparison}
\centering
\scriptsize
\setlength{\tabcolsep}{0.5pt}\renewcommand{\arraystretch}{1.02}
\renewcommand{\tabularxcolumn}[1]{m{#1}}
\begin{tabularx}{\columnwidth}{@{}>{\raggedright\arraybackslash}m{0.18\columnwidth}>{\raggedright\arraybackslash}m{0.17\columnwidth}*{6}{>{\centering\arraybackslash}X}@{}}
\toprule
& & \multicolumn{2}{c}{\shortstack{\textbf{Door-Closing SR}\\\textbf{(\%) $\uparrow$}}} & \multicolumn{2}{c}{\shortstack{\textbf{Survival}\\\textbf{(\%) $\uparrow$}}} & \multicolumn{2}{c}{\shortstack{\textbf{Execution time}\\\textbf{(s) $\downarrow$}}} \\
\cmidrule(lr){3-4}\cmidrule(lr){5-6}\cmidrule(lr){7-8}
\textbf{Method} & \textbf{Interface} & \textbf{Train} & \textbf{Test} & \textbf{Train} & \textbf{Test} & \textbf{Train} & \textbf{Test} \\
\midrule
SONIC & Latent token & 62.10 & 58.30 & 98.53 & 98.49 & 6.25 & 6.27 \\
Handoff +\newline Qwen3-VL-8B & Task-space & 23.81 & 20.94 & 74.82 & 75.09 & 99.98 & 116.89 \\
Handoff +\newline Qwen3-VL-4B & Task-space & 22.54 & 16.79 & 68.07 & 70.38 & 65.46 & 68.86 \\
\cmidrule(lr){1-8}
\textbf{ViLoMan} & \textbf{Joint-level} & \textbf{77.16} & \textbf{77.55} & \textbf{99.22} & \textbf{99.81} & \textbf{5.79} & \textbf{6.20} \\
\bottomrule
\end{tabularx}
\end{table}

\subsection{Effectiveness of Execution Interface}
To address Q2, we compare ViLoMan with two methods that rely on intermediate execution interfaces: SONIC~\cite{luo2025sonic}, which uses a latent-token interface, and Handoff~\cite{yang2026handoff}, which uses an explicit task-space interface. 

For \textbf{SONIC}~\cite{luo2025sonic}, we adapt its universal-token interface. Depth observations and proprioception are mapped to a 64-D token, reshaped to $2\times32$, quantized by its frozen finite scalar quantization (FSQ) module, and decoded by the frozen SONIC decoder into a 29-D joint action conditioned on a ten-frame proprioceptive history. Training uses a squared-action DAgger loss on normalized actions, together with a smooth L1 token loss weighted by 0.1. 
For \textbf{Handoff}~\cite{yang2026handoff}, the official release does not provide its vision-language model (VLM) planner or prompts. We therefore use Qwen3-VL-8B and Qwen3-VL-4B~\cite{bai2025qwen3vl} as high-level planners, with prompts designed for our door-closing task. Given RGB-D observations, each planner generates a 10-dimensional task-space command consisting of planar base motion $(v_x,v_y,\omega_z)$, root height, and 3-D target positions for both wrists. The command is updated at 1~Hz and executed by the Handoff whole-body controller. 

As shown in Table~\ref{tab:handoff_comparison}, ViLoMan achieves a Door-Closing SR of 77.55\% on the test set, substantially outperforming SONIC (58.30\%) and Handoff variants based on Qwen3-VL-8B (20.94\%) and Qwen3-VL-4B (16.79\%). Although SONIC achieves survival rates and execution times comparable to those of ViLoMan, its Door-Closing SR is 19.25 percentage points lower. Handoff exhibits both lower success rates and substantially longer execution times. 
Overall, ViLoMan's direct joint-level interface achieves higher task success than the intermediate-interface baselines while maintaining execution efficiency. Compared with Handoff, it also avoids the substantial inference overhead of VLM-based planning.

\begingroup
\setlength{\intextsep}{5pt}
\setlength{\textfloatsep}{5pt plus 2pt minus 2pt}
\subsection{Closed-Loop Generalization and Real-World Transfer}

\begin{figure}[!t]
    \centering
    \includegraphics[width=0.82\columnwidth]{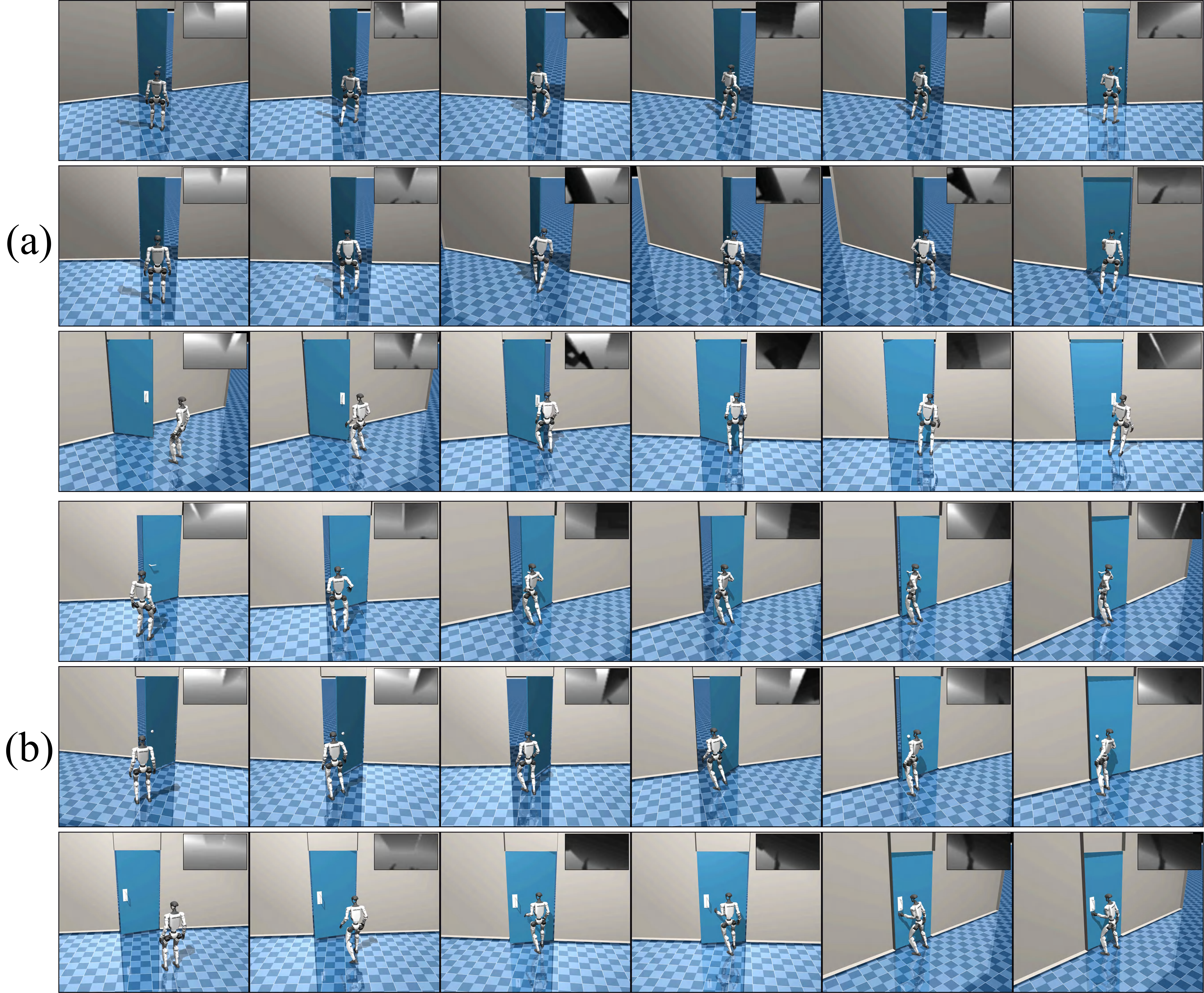}
    \caption{Sim-to-sim evaluation on test articulated-door configurations: (a) left-hinged doors and (b) right-hinged doors. Each row shows a closed-loop rollout, with the inset depth observations as input to the student policy.}
    \label{fig:sim2sim}
\end{figure}

To address Q3, we evaluate ViLoMan's closed-loop adaptation and its direct transfer from simulation to a physical robot. Fig.~\ref{fig:sim2sim} shows sim-to-sim rollouts of the student policy exported in the Open Neural Network Exchange (ONNX) format and deployed in the MuJoCo simulator. Across randomized  initial poses, viewpoints, hinge sides, and door states, the policy uses only depth observations and proprioception to adapt its approach, contact posture, and door-closing motion. This behavior demonstrates that the policy responds to the current robot--door configuration rather than just replaying a fixed trajectory.

We directly deploy the same policy on a physical Unitree G1 robot. The policy runs entirely on the G1's onboard computing unit and receives egocentric depth observations from a head-mounted Intel RealSense D435i camera, together with robot proprioception. The evaluation covers eight combinations of initial door angles and robot poses, with five trials per configuration. As shown in Fig.~\ref{fig:real}, ViLoMan succeeds in 32/40 (80\%) trials, compared with 15/40 (37.5\%) for the BC-only policy and 27/40 (67.5\%) for the policy trained with one quarter of the environments. The successful real-world rollouts take \(6.6\,\mathrm{s}\) on average. These results highlight the importance of online DAgger and sufficient training diversity for real-world transfer.

\begin{figure}[H]
    \centering
    \includegraphics[width=0.82\columnwidth]{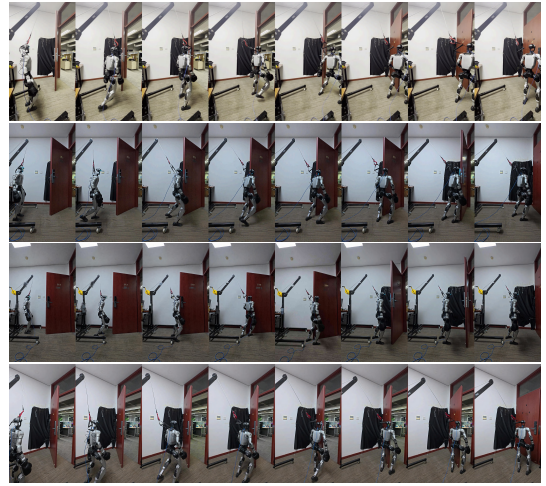}
    \caption{Representative real-world closed-loop rollouts of a Unitree G1 approaching and closing a physical door.}
    \label{fig:real}
\end{figure}

\begin{figure}[!b]
    \centering
    \includegraphics[width=0.82\columnwidth]{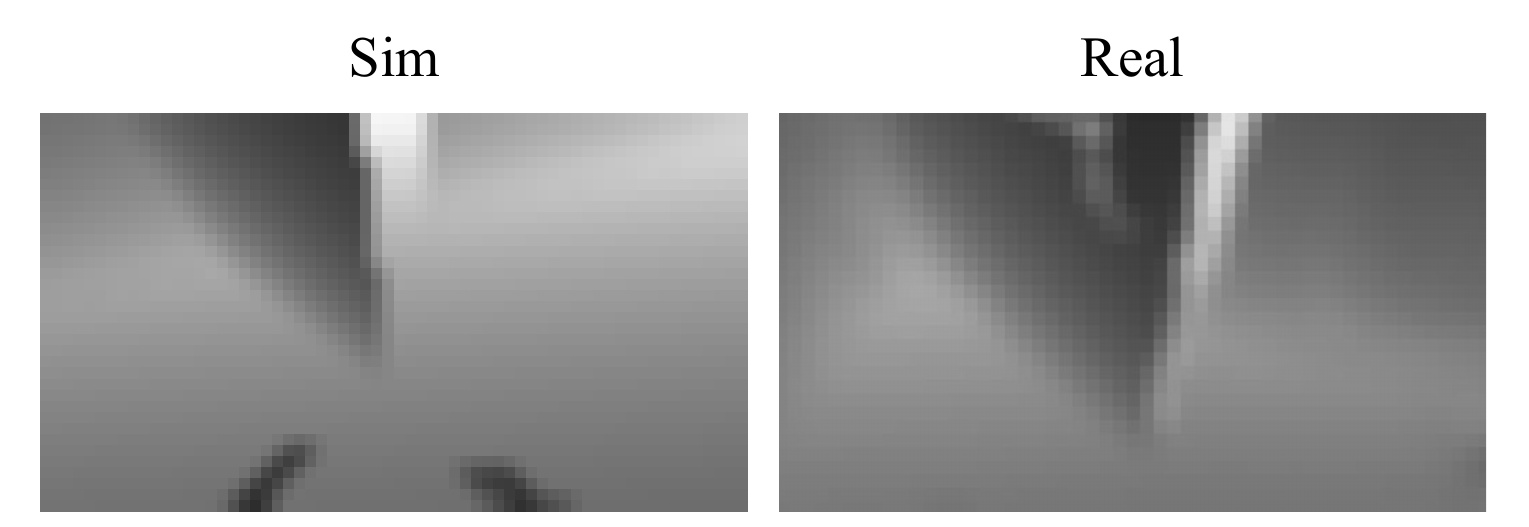}
    \caption{Representative egocentric depth observations from simulation and the real world, as used by the visual--proprioceptive student policy.}
    \label{fig:depth_compare}
\end{figure}

\begin{figure}[!b]
    \centering
    \includegraphics[width=0.84\columnwidth]{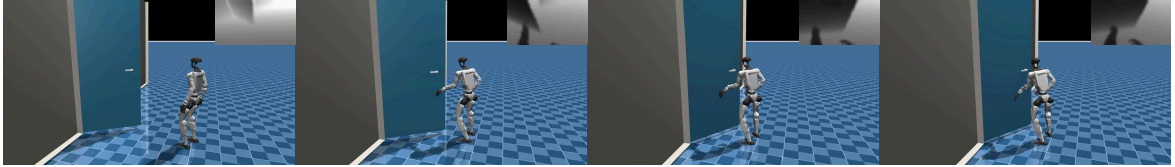}
    \caption{Representative failure case near full door closure.}
    \label{fig:failure_case}
\end{figure}

Fig.~\ref{fig:depth_compare} illustrates egocentric depth observations from simulation and real-world deployment. Despite sensor noise, both domains preserve key geometric cues, including the door panel, door edge, and robot--door configuration, allowing the policy to leverage the spatial representations learned in simulation.
Fig.~\ref{fig:failure_case} shows a representative failure case in which the robot successfully approaches and actuates the door but stops slightly over the \(10^\circ\) threshold, likely due to reduced depth contrast near full closure.

\section{CONCLUSION}

This paper presented \textbf{ViLoMan}, a scalable framework for autonomous humanoid loco-manipulation. By transforming partial kinematic human--object interaction demonstrations into complete, physically executable robot trajectories, ViLoMan enables the learning of a unified whole-body policy through teacher--student distillation. The resulting policy directly maps egocentric depth observations and proprioception to joint-level actions, without relying on reference motions or intermediate commands during deployment. Experiments on articulated-door closing demonstrate that ViLoMan generalizes across diverse task configurations and transfers effectively from simulation to a physical Unitree G1. These results highlight the potential of combining human demonstrations, physics-aware trajectory generation, and visuomotor policy distillation to enable robust and autonomous humanoid loco-manipulation in real-world environments.

\endgroup
\bibliographystyle{IEEEtran}
\bibliography{References}

@article{schulman2017ppo,
  author = {Schulman, John and Wolski, Filip and Dhariwal, Prafulla and Radford, Alec and Klimov, Oleg},
  title = {Proximal Policy Optimization Algorithms},
  journal = {arXiv preprint arXiv:1707.06347},
  year = {2017},
  url = {https://arxiv.org/abs/1707.06347}
}

@inproceedings{ross2011dagger,
  author = {Ross, St{\\'e}phane and Gordon, Geoffrey J. and Bagnell, J. Andrew},
  title = {A Reduction of Imitation Learning and Structured Prediction to No-Regret Online Learning},
  booktitle = {Proceedings of the 14th International Conference on Artificial Intelligence and Statistics},
  pages = {627--635},
  year = {2011}
}

@inproceedings{jiang2024trumans,
  title={Scaling Up Dynamic Human-Scene Interaction Modeling},
  author={Nan Jiang and Zhiyuan Zhang and Hongjie Li and Xiaoxuan Ma and Zan Wang and Yixin Chen and Tengyu Liu and Yixin Zhu and Siyuan Huang},
  booktitle={Proceedings of the IEEE/CVF Conference on Computer Vision and Pattern Recognition (CVPR)},
  pages={1737--1747},
  year={2024}
}

@article{luo2025sonic,
  title={{SONIC}: Supersizing Motion Tracking for Natural Humanoid Whole-Body Control},
  author={Zhengyi Luo and Ye Yuan and Tingwu Wang and Chenran Li and Fernando Castañeda and Sirui Chen and Zi-Ang Cao and Jiefeng Li and David Minor and Qingwei Ben and Jinhyung Park and David Sami and Zi Wang and Xingye Da and Runyu Ding and Cyrus Hogg and Lina Song and Edy Lim and Eugene Jeong and Tairan He and Haoru Xue and Wenli Xiao and Simon Yuen and Jan Kautz and Yan Chang and Umar Iqbal and Linxi "Jim" Fan and Yuke Zhu},
  journal={Science Robotics},
  volume={11},
  number={117},
  pages={eaed4592},
  year={2026},
  doi={10.1126/scirobotics.aed4592},
  url={https://doi.org/10.1126/scirobotics.aed4592}
}

@inproceedings{yang2025omniretarget,
      title={{OmniRetarget}: Interaction-Preserving Data Generation for Humanoid Whole-Body Loco-Manipulation and Scene Interaction}, 
      author={Lujie Yang and Xiaoyu Huang and Zhen Wu and Angjoo Kanazawa and Pieter Abbeel and Carmelo Sferrazza and C. Karen Liu and Rocky Duan and Guanya Shi},
      year={2026},
      eprint={2509.26633},
      booktitle={Proceedings of the IEEE International Conference on Robotics and Automation (ICRA)},
      url={https://arxiv.org/abs/2509.26633}, 
}

@misc{weng2025hdmi,
      title={{HDMI}: Learning Interactive Humanoid Whole-Body Control from Human Videos}, 
      author={Haoyang Weng and Yitang Li and Nikhil Sobanbabu and Zihan Wang and Zhengyi Luo and Tairan He and Deva Ramanan and Guanya Shi},
      year={2025},
      eprint={2509.16757},
      archivePrefix={arXiv},
      primaryClass={cs.RO},
      url={https://arxiv.org/abs/2509.16757}, 
}

@misc{wu2026sugar,
      title={{SUGAR}: A Scalable Human-Video-Driven Generalizable Humanoid Loco-Manipulation Learning Framework}, 
      author={Tianshu Wu and Xiangqi Kong and Yue Chen and Qize Yu and Hang Ye and Jia Li and Yizhou Wang and Hao Dong},
      year={2026},
      eprint={2605.20373},
      archivePrefix={arXiv},
      primaryClass={cs.RO},
      url={https://arxiv.org/abs/2605.20373}, 
}

@misc{xie2026grail,
      title={{GRAIL}: Generating Humanoid Loco-Manipulation from {3D} Assets and Video Priors}, 
      author={Tianyi Xie and Haotian Zhang and Jinhyung Park and Zi Wang and Bowen Wen and Jiefeng Li and Xueting Li and Qingwei Ben and Haoyang Weng and Yufei Ye and David Minor and Tingwu Wang and Chenfanfu Jiang and Sanja Fidler and Jan Kautz and Linxi Fan and Yuke Zhu and Zhengyi Luo and Umar Iqbal and Ye Yuan},
      year={2026},
      eprint={2606.05160},
      archivePrefix={arXiv},
      primaryClass={cs.RO},
      url={https://arxiv.org/abs/2606.05160}, 
}

@misc{rempe2026kimodo,
      title={{Kimodo}: Scaling Controllable Human Motion Generation}, 
      author={Davis Rempe and Mathis Petrovich and Ye Yuan and Haotian Zhang and Xue Bin Peng and Yifeng Jiang and Tingwu Wang and Umar Iqbal and David Minor and Michael de Ruyter and Jiefeng Li and Chen Tessler and Edy Lim and Eugene Jeong and Sam Wu and Ehsan Hassani and Michael Huang and Jin-Bey Yu and Chaeyeon Chung and Lina Song and Olivier Dionne and Jan Kautz and Simon Yuen and Sanja Fidler},
      year={2026},
      eprint={2603.15546},
      archivePrefix={arXiv},
      primaryClass={cs.CV},
      url={https://arxiv.org/abs/2603.15546}, 
}

@misc{zhao2025resmimic,
      title={{ResMimic}: From General Motion Tracking to Humanoid Whole-body Loco-Manipulation via Residual Learning}, 
      author={Siheng Zhao and Yanjie Ze and Yue Wang and C. Karen Liu and Pieter Abbeel and Guanya Shi and Rocky Duan},
      year={2025},
      eprint={2510.05070},
      archivePrefix={arXiv},
      primaryClass={cs.RO},
      url={https://arxiv.org/abs/2510.05070}, 
}

@misc{yin2025visualmimic,
      title={{VisualMimic}: Visual Humanoid Loco-Manipulation via Motion Tracking and Generation}, 
      author={Shaofeng Yin and Yanjie Ze and Hong-Xing Yu and C. Karen Liu and Jiajun Wu},
      year={2025},
      eprint={2509.20322},
      archivePrefix={arXiv},
      primaryClass={cs.RO},
      url={https://arxiv.org/abs/2509.20322}, 
}

@inproceedings{xue2025doorman,
      title={Opening the Sim-to-Real Door for Humanoid Pixel-to-Action Policy Transfer}, 
      author={Haoru Xue and Tairan He and Zi Wang and Qingwei Ben and Wenli Xiao and Zhengyi Luo and Xingye Da and Fernando Castañeda and Guanya Shi and Shankar Sastry and Linxi "Jim" Fan and Yuke Zhu},
      year={2026},
      eprint={2512.01061},
      booktitle={Proceedings of the IEEE/CVF Conference on Computer Vision and Pattern Recognition (CVPR)},
      pages={6642--6652},
      url={https://arxiv.org/abs/2512.01061}, 
}

@inproceedings{he2026viral,
      title={{VIRAL}: Visual Sim-to-Real at Scale for Humanoid Loco-Manipulation}, 
      author={Tairan He and Zi Wang and Haoru Xue and Qingwei Ben and Zhengyi Luo and Wenli Xiao and Ye Yuan and Xingye Da and Fernando Castañeda and Shankar Sastry and Changliu Liu and Guanya Shi and Linxi Fan and Yuke Zhu},
      year={2026},
      eprint={2511.15200},
      booktitle={Proceedings of the IEEE/CVF Conference on Computer Vision and Pattern Recognition (CVPR)},
      pages={13430--13441},
      url={https://arxiv.org/abs/2511.15200}, 
}

@misc{li2026vaic,
      title={{VAIC}: Vision-Guided Humanoid Agile Object Interaction Control via Decoupled Commands}, 
      author={Dongting Li and Qianyang Wu and Xingyu Chen and Liang Li and Yuhang Lin and Sikai Wu and Guoyao Zhang and Mingliang Zhou and Diyun Xiang and Qiang Zhang and Renjing Xu and Jianzhu Ma},
      year={2026},
      eprint={2606.09286},
      archivePrefix={arXiv},
      primaryClass={cs.RO},
      url={https://arxiv.org/abs/2606.09286}, 
}

@misc{gentlehumanoid2025,
      title={{GentleHumanoid}: Learning Upper-body Compliance for Contact-rich Human and Object Interaction}, 
      author={Qingzhou Lu and Yao Feng and Baiyu Shi and Michael Piseno and Zhenan Bao and C. Karen Liu},
      year={2025},
      eprint={2511.04679},
      archivePrefix={arXiv},
      primaryClass={cs.RO},
      url={https://arxiv.org/abs/2511.04679}, 
}

@misc{imagine2real2026,
      title={{Imagine2Real}: Towards Zero-shot Humanoid-Object Interaction via Video Generative Priors}, 
      author={Jiahe Chen and ZiRui Wang and Feiyu Jia and Xiao Chen and Xiaojie Niu and Weishuai Zeng and Tianfan Xue and Xiaowei Zhou and Jiangmiao Pang and Jingbo Wang},
      year={2026},
      eprint={2605.22272},
      archivePrefix={arXiv},
      primaryClass={cs.RO},
      url={https://arxiv.org/abs/2605.22272}, 
}

@misc{yu2026omnicontact,
      title={{OmniContact}: Chaining Meta-Skills via Contact Flow for Generalizable Humanoid Loco-Manipulation}, 
      author={Runyi Yu and Xiaoyi Lin and Ji Ma and Yinhuai Wang and Koukou Luo and Jiahao Ji and Huayi Wang and Wenjia Wang and Runhan Zhang and Ping Tan and Ting Wu and Ruoli Dai and Qifeng Chen and Lei Han},
      year={2026},
      eprint={2606.26201},
      archivePrefix={arXiv},
      primaryClass={cs.RO},
      url={https://arxiv.org/abs/2606.26201}, 
}

@inproceedings{ultra2026,
  title={{ULTRA}: Unified Multimodal Control for Autonomous Humanoid Whole-Body Loco-Manipulation},
  author={Xialin He and Sirui Xu and Xinyao Li and Runpei Dong and Liuyu Bian and Yu-Xiong Wang and Liang-Yan Gui},
  booktitle={IEEE/RSJ International Conference on Intelligent Robots and Systems (IROS)},
  year={2026}
}

@misc{chen2026scenebot,
      title={{SceneBot}: Contact-Prompted General Humanoid Whole Body Tracking with Scene-Interaction}, 
      author={Sirui Chen and Shibo Zhao and Zhen Wu and Jiaman Li and Guanya Shi and C. Karen Liu},
      year={2026},
      eprint={2606.27581},
      archivePrefix={arXiv},
      primaryClass={cs.RO},
      url={https://arxiv.org/abs/2606.27581}, 
}

@inproceedings{zhang2025falcon,
  title={{FALCON}: Learning Force-Adaptive Humanoid Loco-Manipulation},
  author={Yuanhang Zhang and Yifu Yuan and Prajwal Gurunath and Ishita Gupta and Shayegan Omidshafiei and Agha-mohammadi, Ali-Akbar and Marcell Vazquez-Chanlatte and Liam Pedersen and Tairan He and Guanya Shi},
  booktitle={Proceedings of the 8th Annual Learning for Dynamics and Control Conference},
  series={Proceedings of Machine Learning Research},
  volume={331},
  pages={265--281},
  publisher={PMLR},
  year={2026}
}

@misc{yang2026handoff,
      title={{HANDOFF}: Humanoid Agentic Task-Space Whole-Body Control via Distilled Complementary Teachers}, 
      author={Lizhi Yang and Junheng Li and Nehar Poddar and Yiling Hou and Gio Huh and Robert Griffin and Georgia Gkioxari and Aaron Ames},
      year={2026},
      eprint={2606.06493},
      archivePrefix={arXiv},
      primaryClass={cs.RO},
      url={https://arxiv.org/abs/2606.06493}, 
}

@misc{lee2025stageact,
  title={{StageACT}: Stage-Conditioned Imitation for Robust Humanoid Door Opening},
  author={Moonyoung Lee and Dong Ki Kim and Jai Krishna Bandi and Max Smith and Aileen Liao and Agha-mohammadi, Ali-Akbar and Shayegan Omidshafiei},
  year={2025},
  eprint={2509.13200},
  archivePrefix={arXiv},
  primaryClass={cs.RO},
  url={https://arxiv.org/abs/2509.13200},
}

@misc{wu2026php,
  title={Perceptive Humanoid Parkour: Chaining Dynamic Human Skills via Motion Matching},
  author={Zhen Wu and Xiaoyu Huang and Lujie Yang and Yuanhang Zhang and Xi Chen and Pieter Abbeel and Rocky Duan and Angjoo Kanazawa and Carmelo Sferrazza and Guanya Shi and C. Karen Liu},
  year={2026},
  eprint={2602.15827},
  archivePrefix={arXiv},
  primaryClass={cs.RO},
  url={https://arxiv.org/abs/2602.15827}
}

@article{li2023omomo,
  title={Object Motion Guided Human Motion Synthesis},
  author={Jiaman Li and Jiajun Wu and C. Karen Liu},
  journal={ACM Transactions on Graphics},
  volume={42},
  number={6, Art. no. 202},
  month=dec,
  year={2023},
  doi={10.1145/3618333}
}

@inproceedings{ze2025twist2,
  title={{TWIST2}: Scalable, Portable, and Holistic Humanoid Data Collection System},
  author={Yanjie Ze and Siheng Zhao and Weizhuo Wang and Angjoo Kanazawa and Rocky Duan and Pieter Abbeel and Guanya Shi and Jiajun Wu and C. Karen Liu},
  booktitle={Proceedings of the IEEE International Conference on Robotics and Automation (ICRA)},
  year={2026},
  eprint={2511.02832},
  archivePrefix={arXiv},
  primaryClass={cs.RO},
  
}

@misc{yu2026oasis,
  title={{OASIS}: From Simulation Data Collection to Real-World Humanoid Loco-Manipulation},
  author={Zehao Yu and Jiakun Zheng and Weiji Xie and Jiyuan Shi and Chenyun Zhang and Chenjia Bai and Xuelong Li},
  year={2026},
  eprint={2606.08548},
  archivePrefix={arXiv},
  primaryClass={cs.RO},
  url={https://arxiv.org/abs/2606.08548}
}

@misc{zhou2026exoactor,
  title={{ExoActor}: Exocentric Video Generation as Generalizable Interactive Humanoid Control},
  author={Yanghao Zhou and Jingyu Ma and Yibo Peng and Zhenguo Sun and Yu Bai and B{\"o}rje F. Karlsson},
  year={2026},
  eprint={2604.27711},
  archivePrefix={arXiv},
  primaryClass={cs.RO},
  url={https://arxiv.org/abs/2604.27711}
}

@inproceedings{pavlakos2019smplx,
  title={Expressive Body Capture: {3D} Hands, Face, and Body from a Single Image},
  author={Pavlakos, Georgios and Choutas, Vasileios and Ghorbani, Nima and Bolkart, Timo and Osman, Ahmed A. A. and Tzionas, Dimitrios and Black, Michael J.},
  booktitle={Proceedings of the IEEE/CVF Conference on Computer Vision and Pattern Recognition},
  pages={10975--10985},
  year={2019}
}

@inproceedings{lu2025humoto,
  title={{HUMOTO}: A {4D} Dataset of Mocap Human Object Interactions},
  author={Lu, Jiaxin and Huang, Chun-Hao Paul and Bhattacharya, Uttaran and Huang, Qixing and Zhou, Yi},
  booktitle={Proceedings of the IEEE/CVF International Conference on Computer Vision (ICCV)},
  pages={10886--10897},
  year={2025}
}

@inproceedings{araujo2025gmr,
  title={Retargeting Matters: General Motion Retargeting for Humanoid Motion Tracking},
  author={Araujo, Joao Pedro and Ze, Yanjie and Xu, Pei and Wu, Jiajun and Liu, C. Karen},
  booktitle={Proceedings of the IEEE International Conference on Robotics and Automation (ICRA)},
  year={2026},
  url={https://arxiv.org/abs/2510.02252}
}

@misc{zakka2026mjlab,
  title={{mjlab}: A Lightweight Framework for {GPU}-Accelerated Robot Learning},
  author={Zakka, Kevin and Liao, Qiayuan and Yi, Brent and Le Lay, Louis and Sreenath, Koushil and Abbeel, Pieter},
  year={2026},
  eprint={2601.22074},
  archivePrefix={arXiv},
  primaryClass={cs.RO},
  url={https://arxiv.org/abs/2601.22074}
}

@misc{urakami2019doorgym,
  title={{DoorGym}: A Scalable Door Opening Environment and Baseline Agent},
  author={Urakami, Yusuke and Hodgkinson, Alec and Carlin, Casey and Leu, Randall and Rigazio, Luca and Abbeel, Pieter},
  year={2019},
  eprint={1908.01887},
  archivePrefix={arXiv},
  primaryClass={cs.RO},
  url={https://arxiv.org/abs/1908.01887}
}

@misc{bai2025qwen3vl,
  title={{Qwen3-VL Technical Report}},
  author={Shuai Bai and Yuxuan Cai and Ruizhe Chen and Keqin Chen and Xionghui Chen and Zesen Cheng and others},
  year={2025},
  eprint={2511.21631},
  archivePrefix={arXiv},
  primaryClass={cs.CV},
  url={https://arxiv.org/abs/2511.21631}
}


\clearpage
\appendix
We detail dataset diversity (Table~\ref{tab:dataset_diversity}), domain
randomization (Table~\ref{tab:domain_randomization}), the expanded teacher
reward (Table~\ref{tab:privileged_teacher_reward}), and training
hyperparameters (Table~\ref{tab:training_hyperparameters}), together with
the student interface and reward notation.

\subsection{Dataset Diversity}
The 409 training and 106 test trajectories use disjoint source TRUMANS
clips. Table~\ref{tab:dataset_diversity} reports the categorical composition
and geometric ranges; percentages are rounded within each split.
\setcounter{table}{3}
\begin{table}[H]
\caption{Training and test dataset diversity.}
\label{tab:dataset_diversity}
\centering
\footnotesize
\setlength{\tabcolsep}{2.2pt}
\renewcommand{\arraystretch}{1.08}
\begin{tabularx}{\columnwidth}{@{}Xcc@{}}
\toprule
\textbf{Diversity attribute} & \textbf{Train split} & \textbf{Test split} \\
\midrule
Number of trajectories & 409 & 106 \\
Handle categories & 3 & 3 \\
Lever handle & 134 (32.8\%) & 35 (33.0\%) \\
Pull handle & 139 (34.0\%) & 36 (34.0\%) \\
Round handle & 136 (33.3\%) & 35 (33.0\%) \\ Left-hinged door & 273 (66.7\%) & 42 (39.6\%) \\ Right-hinged door & 136 (33.3\%) & 64 (60.4\%) \\
Initial door opening angle & $10.05^\circ$--$80.31^\circ$ & $22.82^\circ$--$83.28^\circ$ \\
Door height & 1.981--2.473 m & 1.986--2.473 m \\
Door width & 1.000 m & 1.000 m \\
Door thickness & 0.020--0.029 m & 0.020--0.029 m \\
Longitudinal distance $d_x$ & 1.201--1.799 m & 1.202--1.799 m \\
Lateral offset $d_y$ & $-0.247$--0.250 m & $-0.248$--0.245 m \\

\bottomrule
\end{tabularx}
\end{table}

\subsection{Observation and Action Interface}
At 50\,Hz, body-frame angular velocity, projected gravity, joint positions,
and joint velocities are sampled at offsets $[0,1,2,3,4,8,12,16,20]$.
Adding eight previous actions gives $9(3+3+29+29)+8\cdot29=808$ features.
Four consecutive $36\times64$ depth frames are normalized using a
2.5\,m maximum depth and encoded into a 32-D feature by a residual CNN.
The 29-D output is scaled and added to the nominal joint configuration.
Student inputs exclude reference motions and privileged object states.

\subsection{Stage Gates and Reward Scope}
Stage~0 precedes the reference walk-end frame, Stage~1 lasts until
closing-start, and Stage~2 follows. We use $g_k=\mathbf{1}[s_t=k]$,
$p_t=t/\max(T-1,1)$ for $T$ reference frames, and $r_t=\sum_i w_i r_{i,t}$.
Table~\ref{tab:privileged_teacher_reward} expands 34 enabled terms of the
reported configuration; specialist coefficients can differ. Separate
plain-survival and foot-contact-count terms are disabled in this configuration.

$C_{f,t}$ and $C^*_{f,t}$ are actual and reference foot-contact indicators;
$F_{f,t}$ indicates first contact. Actual contact uses at least half the
physics-substep votes. The timer $u_{f,t}$ starts at zero on reset.
$h_f^{\min}$ and $h_f^{\max}$ are the minimum and maximum ankle-roll/toe
heights. $\overline{\dot q}_{j,t}$ averages the two latest physics-substep
velocity buffers. Soft position limits retain 90\% of the joint range;
soft torque limits use $0.75\tau_j^{\max}$. The acceleration penalty uses
a one-sided upper clamp at 100\,rad/s$^2$.

\newpage
\subsection{Domain Randomization}
Table~\ref{tab:domain_randomization} reports the sampling ranges. $\mathcal U$
and $\mathcal{LU}$ denote uniform and log-uniform distributions.
Multipliers scale nominal parameters; offsets use the stated units.
\setcounter{table}{5}
\begin{table}[H]
\caption{Domain-randomization settings.}
\label{tab:domain_randomization}
\centering
\footnotesize
\setlength{\tabcolsep}{2pt}
\renewcommand{\arraystretch}{1.00}
\begin{tabularx}{\columnwidth}{@{}>{\raggedright\arraybackslash}p{0.52\columnwidth}>{\raggedright\arraybackslash}X@{}}
\toprule
\textbf{Parameter} & \textbf{Sampling distribution} \\
\midrule
\multicolumn{2}{@{}l}{\textit{Robot dynamics}} \\
Pelvis / torso center-of-mass (CoM) shift & $\Delta c_{x,y,z}\sim\mathcal{U}[-0.03,0.03]$ m \\
Upper-body stiffness/damping scale & $\mathcal{LU}[0.9,1.1]$ \\
Lower-body/waist stiffness/damping scale & $\mathcal{LU}[0.5,2.0]$ \\
Armature multiplier & $\mathcal{LU}[0.75,1.25]$ \\
Joint-position offset & $\Delta q\sim\mathcal{U}[-0.01,0.01]$ rad \\
Lower-body target joint-position bias & $\Delta q_{\mathrm{target}}\sim\mathcal{U}[-0.1,0.1]$ rad \\
\midrule
\multicolumn{2}{@{}l}{\textit{Contact and ground}} \\
Foot sliding friction coefficient & $\mu_s\sim\mathcal{U}[0.3,1.2]$ \\
Contact time constant & $\mathcal{U}[0.015,0.03]$ s \\
Contact damping ratio & $\mathcal{LU}[0.5,2.0]$ \\
\midrule
\multicolumn{2}{@{}l}{\textit{Door dynamics}} \\
Hinge damping multiplier & $\mathcal{U}[0.9,1.1]$ \\
Hinge friction multiplier & $\mathcal{U}[0.9,1.2]$ \\
Hinge armature multiplier & $\mathcal{U}[0.9,1.1]$ \\
\midrule
\multicolumn{2}{@{}l}{\textit{External perturbations}} \\
Reference-root linear-velocity drift in $x,y$ & $\|\mathbf v_{xy}\|\sim\mathcal{U}[0,0.25]$ m/s \\
Reference-root linear-velocity drift in $z$ & $v_z\sim\mathcal{U}[-0.05,0.05]$ m/s \\
Root-height offset & $\mathcal{U}[-0.03,0.03]$ m \\
Perturbation period & $\mathcal{U}[4,6]$ s \\
Linear push velocity & $v_{x,y}\sim\mathcal{U}[-0.5,0.5]$, $v_z\sim\mathcal{U}[-0.2,0.2]$ m/s \\
Angular push velocity & $\omega_{\mathrm{roll,pitch}}\sim\mathcal{U}[-0.52,0.52]$, $\omega_{\mathrm{yaw}}\sim\mathcal{U}[-0.78,0.78]$ rad/s \\
Gravity perturbation & $\boldsymbol{\epsilon}_g=r\mathbf u$, $r\sim\mathcal{U}[0,0.1]\,\mathrm{m/s^2}$, $\mathbf u\sim\operatorname{Unif}(\mathbb S^2)$ \\
\bottomrule
\end{tabularx}
\end{table}

\subsection{Reward Notation}
An asterisk denotes a reference value; superscripts $w$, $b$, $xy$, and
$\mathrm{rel}$ denote world-frame, body-frame, horizontal-plane, and
root-relative quantities. $K$, $K_l$, and $K_u$ are the complete, lower-body,
and upper-body keypoint sets; $J$ is the selected non-ankle joint set.
The targets $\tilde p_i^{w,*}$ align the reference root pose with the current
root, whereas Stage~0 global keypoint tracking uses unaligned world targets.
Rotation error is $d_R(R_1,R_2)=\|\operatorname{axisangle}(R_1R_2^{-1})\|_2$.
Root velocities are expressed in their respective body frames; relative
keypoint velocities additionally subtract root velocity.

$c_h^*$ identifies the reference interaction hand, $d_h$ is its distance
to the current door-panel rectangle, and $s_h$ is signed distance to the
door plane. These geometric terms do not measure contact force.
$\theta_0^*$ and $\theta_f^*$ denote the initial and final reference hinge
angles, with $\Delta\theta^*=\theta_f^*-\theta_0^*$.
$v_t$ is the validity flag for a previous Stage~0 XY error, and is zero
on the first update after reset or stage entry.

\clearpage
\makeatletter
\setlength{\@fptop}{0pt}
\setlength{\@dblfptop}{0pt}
\makeatother

\begin{table*}[!t]
\setcounter{table}{4}
\caption{Expanded reward specification for a representative interaction-teacher configuration. Expressions are unweighted.}
\label{tab:privileged_teacher_reward}
\centering
\begingroup
\fontsize{7.4}{8.2}\selectfont
\setlength{\tabcolsep}{3pt}
\renewcommand{\arraystretch}{0.90}
\noindent $\rho_{\Sigma}(e)=|\Sigma|^{-1}\sum_{\sigma\in\Sigma}\exp(-e/\sigma)$,
with positive decay scales in the units of $e$; $\Delta t=0.02$\,s.
\par\vspace{2pt}
\begin{tabularx}{\textwidth}{@{}>{\raggedright\arraybackslash}p{0.22\textwidth}>{\raggedright\arraybackslash}Xrc@{}}
\toprule
\textbf{Reward term} & \textbf{Unweighted expression $r_{i,t}$} & \textbf{Weight $w_i$} & \textbf{Stage} \\
\midrule
\rowcolor{gray!10}\multicolumn{4}{@{}l}{\textit{Stage 0: walk and approach}} \\
Root XY tracking & $g_0\,\rho_{\{0.30,1.00\}}(e_t^{xy}),\quad e_t^{xy}=\|p_{r,t}^{xy}-p_{r,t}^{xy,*}\|_2$ & $2.00$ & 0 \\
Root XY progress & $g_0\,v_t\operatorname{clip}((e_{t-1}^{xy}-e_t^{xy})/\Delta t,-1,1)$ & $1.00$ & 0 \\
Global lower-body keypoints & $g_0\,\rho_{\{0.30,1.00\}}\!(\tfrac{1}{|K_l|}\sum_{i\in K_l}\|p_{i,t}^w-p_{i,t}^{w,*}\|_2)$ & $1.00$ & 0 \\
\midrule
\rowcolor{gray!10}\multicolumn{4}{@{}l}{\textit{Stage 1: pre-contact stabilization}} \\
End-effector pre-contact tracking & $g_1\,\operatorname{mean}_h\![(1-c_h^*)+2c_h^*\exp\!(-\tfrac{\max(d_h-0.06,0)}{0.18})]$ & $1.00$ & 1 \\
\midrule
\rowcolor{gray!10}\multicolumn{4}{@{}l}{\textit{Stage 2: door closing and contact}} \\
Hinge-angle tracking & $g_2\,\rho_{\{0.80,0.30,0.10\}}(|\theta_t-\theta_t^*|)$ & $18.0$ & 2 \\
Hinge-angle $L_1$ penalty & $-g_2\,|\theta_t-\theta_t^*|$ & $4.00$ & 2 \\
Final hinge-angle tracking & $g_2\cdot\mathbf{1}[p_t\ge0.35]\cdot\rho_{\{0.35,0.12\}}(|\theta_t-\theta_f^*|)$ & $12.0$ & 2 \\
Final hinge-angle penalty & $-g_2\cdot\mathbf{1}[p_t\ge0.35]\cdot|\theta_t-\theta_f^*|$ & $8.00$ & 2 \\
Directed hinge progress & $g_2\cdot\mathbf{1}[p_t\ge0.25]\cdot\mathbf{1}[|\Delta\theta^*|\ge0.05]\cdot\operatorname{clip}\!(\tfrac{(\theta_t-\theta_0^*)\operatorname{sgn}(\Delta\theta^*)}{\max(|\Delta\theta^*|,10^{-6})},0,1)$ & $6.00$ & 2 \\
Hinge-velocity tracking & $g_2\,\rho_{\{1.0\}}(|\dot\theta_t-\dot\theta_t^*|)$ & $0.20$ & 2 \\
End-effector contact tracking & $g_2\,\operatorname{mean}_h\![(1-c_h^*)+4c_h^*\exp\!(-\tfrac{\max(d_h-0.04,0)}{0.12})]$ & $2.80$ & 2 \\
Contact-plane penetration & $-g_2\,\operatorname{mean}_h[c_h^*\max(-0.02-s_h,0)]$ & $0.45$ & 2 \\
\midrule
\rowcolor{gray!10}\multicolumn{4}{@{}l}{\textit{Across stages: survival}} \\
Progress-gated survival & $g_0\exp(-e_t^{xy}/1.0)+g_1+g_2$ & $2.00$ & 0--2 \\
\midrule
\rowcolor{gray!10}\multicolumn{4}{@{}l}{\textit{Always-on motion tracking}} \\
Root position & $\rho_{\{0.30\}}(\|p_{r,t}^w-p_{r,t}^{w,*}\|_2)$ & $0.50$ & 0--2 \\
Root orientation & $\rho_{\{0.40\}}(d_R(R_{r,t}^w,R_{r,t}^{w,*}))$ & $0.50$ & 0--2 \\
Root linear velocity & $\rho_{\{1.00\}}(\|v_{r,t}^b-v_{r,t}^{b,*}\|_2)$ & $1.00$ & 0--2 \\
Root angular velocity & $\rho_{\{3.00\}}(\|\omega_{r,t}^b-\omega_{r,t}^{b,*}\|_2)$ & $1.00$ & 0--2 \\
All keypoint positions & $\rho_{\{0.30\}}\!(\operatorname{mean}_{i\in K}\|p_{i,t}^w-\tilde p_{i,t}^{w,*}\|_2)$ & $1.00$ & 0--2 \\
Keypoint linear velocities & $\rho_{\{1.00\}}\!(\operatorname{mean}_{i\in K}\|v_{i,t}^{b,\mathrm{rel}}-v_{i,t}^{b,\mathrm{rel},*}\|_2)$ & $1.00$ & 0--2 \\
Keypoint orientations & $\rho_{\{0.40\}}\!(\operatorname{mean}_{i\in K}d_R(R_{i,t}^b,R_{i,t}^{b,*}))$ & $1.00$ & 0--2 \\
Keypoint angular velocities & $\rho_{\{3.00\}}\!(\operatorname{mean}_{i\in K}\|\omega_{i,t}^{b,\mathrm{rel}}-\omega_{i,t}^{b,\mathrm{rel},*}\|_2)$ & $1.00$ & 0--2 \\
Lower-body keypoint positions & $\rho_{\{0.20\}}\!(\operatorname{mean}_{i\in K_l}\|p_{i,t}^w-\tilde p_{i,t}^{w,*}\|_2)$ & $0.50$ & 0--2 \\
Upper-body keypoint positions & $\rho_{\{0.30\}}\!(\operatorname{mean}_{i\in K_u}\|p_{i,t}^w-\tilde p_{i,t}^{w,*}\|_2)$ & $0.35$ & 0--2 \\
Joint positions & $\rho_{\{0.50\}}(\operatorname{mean}_{j\in J}|q_{j,t}-q_{j,t}^*|)$ & $0.75$ & 0--2 \\
Joint velocities & $\rho_{\{3.00\}}(\operatorname{mean}_{j\in J}|\dot q_{j,t}-\dot q_{j,t}^*|)$ & $0.50$ & 0--2 \\
\midrule
\rowcolor{gray!10}\multicolumn{4}{@{}l}{\textit{Always-on regularization}} \\
Residual action magnitude & $-\tfrac1{29}\sum_{j=1}^{29}(\Delta a_{j,t}^{\mathrm{res}})^2$ & $2\!\times\!10^{-4}$ & 0--2 \\
Residual action change & $-\tfrac1{29}\sum_{j=1}^{29}(\Delta a_{j,t}^{\mathrm{res}}-\Delta a_{j,t-1}^{\mathrm{res}})^2$ & $5\!\times\!10^{-4}$ & 0--2 \\
Joint velocity penalty & $-\sum_j(\overline{\dot q}_{j,t})^2$ & $5\!\times\!10^{-4}$ & 0--2 \\
Total action change & $-\sum_j(a_{j,t}-a_{j,t-1})^2$ & $0.01$ & 0--2 \\
Reference-aware foot timing & $\begin{aligned}u^-_{f,t}&=u_{f,t-1}+\Delta t(1-2\mathbf{1}[C_{f,t}\ne C^*_{f,t}]),\\[-.6ex]r_{i,t}&=\sum_f\min(u^-_{f,t}-0.5,0)F_{f,t},\quad u_{f,t}=u^-_{f,t}(1-C_{f,t})\end{aligned}$ & $5.00$ & 0--2 \\
Dense foot contact and clearance & $\operatorname{mean}_f\!\begin{cases}-1,&C_f\ne C_f^*,\\[-.6ex]-1+\operatorname{clip}((h_f^{\min}-0.035)/0.120,0,1),&C_f=C_f^*=0,\\[-.6ex]-\operatorname{clip}((h_f^{\max}-0.035)/0.090,0,1),&C_f=C_f^*=1.\end{cases}$ & $1.00$ & 0--2 \\
Soft joint-position limits & $-\tfrac{\sum_j(\max(q_j^{\mathrm{lo,soft}}-q_j,0)+\max(q_j-q_j^{\mathrm{hi,soft}},0))}{1-0.9}$ & $1.00$ & 0--2 \\
Soft actuator-torque limits & $-\sum_j\max(|\tau_j|/(0.75\tau_j^{\max})-1,0)$ & $0.01$ & 0--2 \\
Joint acceleration penalty & $-\sum_j\min(\ddot q_j,100)^2$ & $5\!\times\!10^{-7}$ & 0--2 \\
\bottomrule
\end{tabularx}
\endgroup

\vspace{1pt}
\setcounter{table}{6}
\caption{Training hyperparameters for the teacher and student pipelines.}
\label{tab:training_hyperparameters}
\begingroup
\fontsize{7.5}{8.4}\selectfont
\setlength{\tabcolsep}{3pt}
\renewcommand{\arraystretch}{1.00}
\begin{tabularx}{\textwidth}{@{}>{\raggedright\arraybackslash}p{0.19\textwidth}>{\raggedright\arraybackslash}X>{\raggedright\arraybackslash}p{0.19\textwidth}>{\raggedright\arraybackslash}X@{}}
\toprule
\textbf{Teacher: PPO} & \textbf{Value} & \textbf{Student: DAgger} & \textbf{Value} \\
\midrule
Actor / critic hidden widths & $1024\to512\to512$ & Policy MLP / activation & $512\to256\to128$ / ELU \\
Optimizer / initial LR & Separate Adam; $10^{-4}$ & Depth encoder / feature & Residual CNN / 32-D \\
Parallel environments / process & 1,024 & Proprioception / depth input & 808-D / $4\times36\times64$ \\
Rollout steps / collection & 32 & Output action & 29-D joint-position action \\
Discount / GAE $\lambda$ & 0.99 / 0.95 & Optimizer / learning rate & AdamW / $10^{-4}$ \\
PPO clip / desired KL & 0.2 / 0.01 & Persistent rollout chunk & 32 steps (0.64\,s); retain state \\
Initial action std. dev. & 0.25 & Minibatch / update epochs & 1,024 / 1 \\
Entropy coefficient & $0.0025\to0.0005$; exponential & Update samples / worker & $98{,}304=65{,}536+32{,}768$ \\
Maximum gradient norm & 1.0, actor and critic separately & Current / replay fraction & $2/3$ / $1/3$ \\
Value loss / coefficient & MSE / 1.0 & Replay capacity / worker & 300,000 samples \\
Control / physics time step & 0.02\,s / 0.005\,s & Replay storage & float16 \\
Stage reset probabilities & 0.25 / 0.15 / 0.60 (Stages 0/1/2) & Action-matching loss / weight & Normalized action MSE / 1.0 \\
Reset start & Corresponding stage boundary & Control frequency & 50\,Hz \\
GMT prior & Frozen & Maximum normalized depth & 2.5\,m \\
\bottomrule
\end{tabularx}
\par\vspace{1pt}
\endgroup
\end{table*}
\clearpage
\end{document}